\documentclass[journal]{IEEEtran}
\usepackage[utf8]{inputenc}
\usepackage{amsmath}
\usepackage{amssymb}
\usepackage{algorithm}
\usepackage{algpseudocode}
\usepackage{csquotes}
\usepackage{array}
\usepackage{graphicx}
\usepackage{caption}
\usepackage{subfig}

\begin{document}
\title{Decentralized Coverage Path Planning with Reinforcement Learning and Dual Guidance}

\author{Yongkai~Liu, Jiawei~Hu, Wei~Dong%
\thanks{Manuscript received;}%
\thanks{Y.~Liu, J.~Hu and W.~Dong are with the State Key Laboratory of Mechanical System and Vibration, School of Mechanical Engineering, Shanghai Jiao Tong University, Shanghai, 200240, China. (Corresponding author: dr.dongwei@sjtu.edu.cn)}}

\maketitle

\begin{abstract}
Planning coverage path for multiple robots in a decentralized way enhances robustness to coverage tasks handling uncertain malfunctions. To achieve high efficiency in a distributed communication system, a comprehensive understanding of both the complicated environments and cooperative agents' intent is crucial. Unfortunately, existing works commonly consider only part of these factors, resulting in imbalanced subareas or unnecessary overlaps. To tackle this issue, we introduce a Decentralized reinforcement learning framework with dual guidance (DODGE) to train each agent to solve the decentralized multiple coverage path planning problem straightly through the environment states. As distributed robots require others' intentions to perform better coverage efficiency, we utilize two guidance methods, artificial potential fields and heuristic guidance, to include and integrate others' intentions into sensored observations for each robot. With our constructed framework, results have shown our agents successfully learn to determine their own subareas while achieving full coverage, balanced subareas and low overlap rates. We then implement spanning tree cover within those subareas to construct actual routes for each robot and complete given coverage tasks. Our performance is also compared with the state-of-the-art decentralized method showing at most 10\% lower overlap rates while performing high efficiency in similar environments. 
\end{abstract}


\section{Introduction} \label{sec:introduction}

\IEEEPARstart{M}{ultiple} coverage path planning (mCPP) has been widely adopted in numerous tasks such as search and rescue \cite{9220149}, floor cleaning \cite{1337300}, boundary inspection \cite{1570204}, industrial contour-following \cite{6304921}, etc. For mCPP, one popular way is to first decompose the area of interest (AOI) while in the offline state and then apply sweeping methods like spanning tree cover (STC) \cite{1013479} or lawnmower \cite{azpurua2018multi} within the subareas of the AOI.
Among the above approaches, one of the main challenges is to allocate balanced subareas for each robot such that the coverage tasks can be completed more efficiently. Moreover, the allocated subareas are desired to have less overlap area so that less congestion will occur to hurt the overall performance.

Some of the previous research, known as centralized methods, have shown complete coverage \cite{5699392} and optimal division \cite{kapoutsisdarp} in the planning phase assuming access to global information and all robots. As relying on a central controller, a malfunction in the central control station or any robot might cause the whole system to break down or missions to be finished incompletely. On the other hand, in decentralized methods, robots take care of their own decisions independently, providing higher fault tolerance than centralized strategies. Among recent decentralized works, Vishnu et al. \cite{nair2020gm} proposed a Geodesic-Manhattan Voronoi-partition coverage (GM-VPC) algorithm that accomplished full coverage of the AOI. Although successfully decomposed the AOI, the efficiency of the planned routes depends on the obstacle distribution and the initial positions of the robots. Aiming to acquire higher efficiency, an artificially weighted spanning tree cover (AWSTC) algorithm \cite{9017928} is proposed to allocate subareas fairly. AWSTC recursively constructs spanning trees, later for STC, in a heuristic way far from other robots and the center of inertia of the uncovered area while avoiding redundant coverage. This method demonstrated higher efficiency than the centralized method, multi-robot forest coverage (MFC) \cite{1545323}, and brought in system resilience introduced by the decentralized re-planner. Despite that, AWSTC greedily extends all robots' spanning trees in the aforementioned heuristic manner resulting in unnecessary overlaps in environments like outdoor-like terrains. Thus far, preserving balanced routes for all robots while minimizing overlaps remains challenging in the decentralized mCPP problem.

As it is hard to model the complicated environments, including obstacles and other robots in decentralized mCPP tasks, existing decentralized methods utilize only part of the factors in the environments resulting in imbalance subareas or unnecessary overlap areas. Motivated by the fact that human beings can directly judge the given area is more likely belonging to which robot by simply looking at the area and the robots' positions. Many researchers have worked on learning the complicated environments instead of straightly modeling them. Lately, a distributed mCPP algorithm named Q-traversal \cite{8963663} based on reinforcement learning has come up and presented better performance than the distributed anti-flock algorithm \cite{1605401} for dynamic area coverage. On the fact that Q-traversal is based on value tables \cite{watkins1992q}, it suffers from the dimension explosion problem making it unsuitable for large map sizes. To alleviate the dimension curse, researchers combine deep learning and map the observation to a series of actions \cite{8103164}. An mCPP algorithm standing on deep reinforcement learning (DRL) framework is proposed to solve patrolling tasks for autonomous surface vehicles on the Ypacaraí Lake \cite{9330612}. While the authors claimed a centralized-distributed deep Q-leaning network, the routes were generated by a central controller. Thus, the aforementioned robustness issue in centralized systems also remains in the algorithm. A decentralized DRL framework capable of resolving mCPP problems is still required in the scope. Despite that, it's more difficult to develop a decentralized DRL framework than a centralized one. Decentralized systems often bring instability to the training phase of a learner policy due to the lack of other robots' intentions \cite{9166753}. Augmenting additional aids into a decentralized DRL framework is crucial for robots to learn better policy.

Inspired by both the spanning method in AWSTC \cite{9017928} and the mapping approach as the Ypacaraí Lake planner \cite{9330612}, we head to construct a decentralized deep reinforcement learning framework that trains robots to determine their subareas by repulsively expanding their territories from their initial cells in a way that is capable of achieving balanced subareas while reducing overlap areas according to the given area and the planned subarea. Additionally, to help learn better policy, we also require some guiding approaches, but instead of using subgoal rewards \cite{9733281} which are hard to define in our coverage tasks, we include two guiding methods to give our robots more information and improve the coverage performance. Furthermore, in order to address the issue mentioned above in AWSTC, a lazy-evaluation action to let robots intelligently decide whether they should hold their decision or keep expanding is also desired in our structure. To verify our method, we focus on the outdoor-like environments where the generated routes should have comparable low redundancy while retaining balanced burdens on all robots.

The main contributions of our work are as follows. 
\begin{itemize}
  \item[1)] Present a thorough reinforcement learning framework that trains robots to distributively allocate their subareas in a way that the later STC-generated routes have sufficiently high efficiency and low overlap rates for mCPP tasks.
  \item[2)] Improve the planning decision of leaner-policy by including two guiding methods, artificial potential fields and heuristic guidance.
  \item[3)] Prevent unnecessary overlaps of the allocated subareas by designing an extra lazy-evaluation action.
\end{itemize}
The remaining content is organized as follows: Section \ref{sec:problem description} describes the problem. Section \ref{sec:RL problem casting} presents the reinforcement learning casting of our mCPP problem. Section \ref{sec:NNpolicy} gives the design of our neural network policy. Section \ref{sec:training} details the training scenarios and options. Section \ref{sec6} shows the results of our experiments including ablation experiments of our framework, extension tests, performance comparison with other algorithms and an on-board experiment. The conclusion is drawn in Section \ref{sec7}.

\begin{figure}[t]
\captionsetup{justification=raggedright,singlelinecheck=false}
\centering
\includegraphics[width=0.48\textwidth]{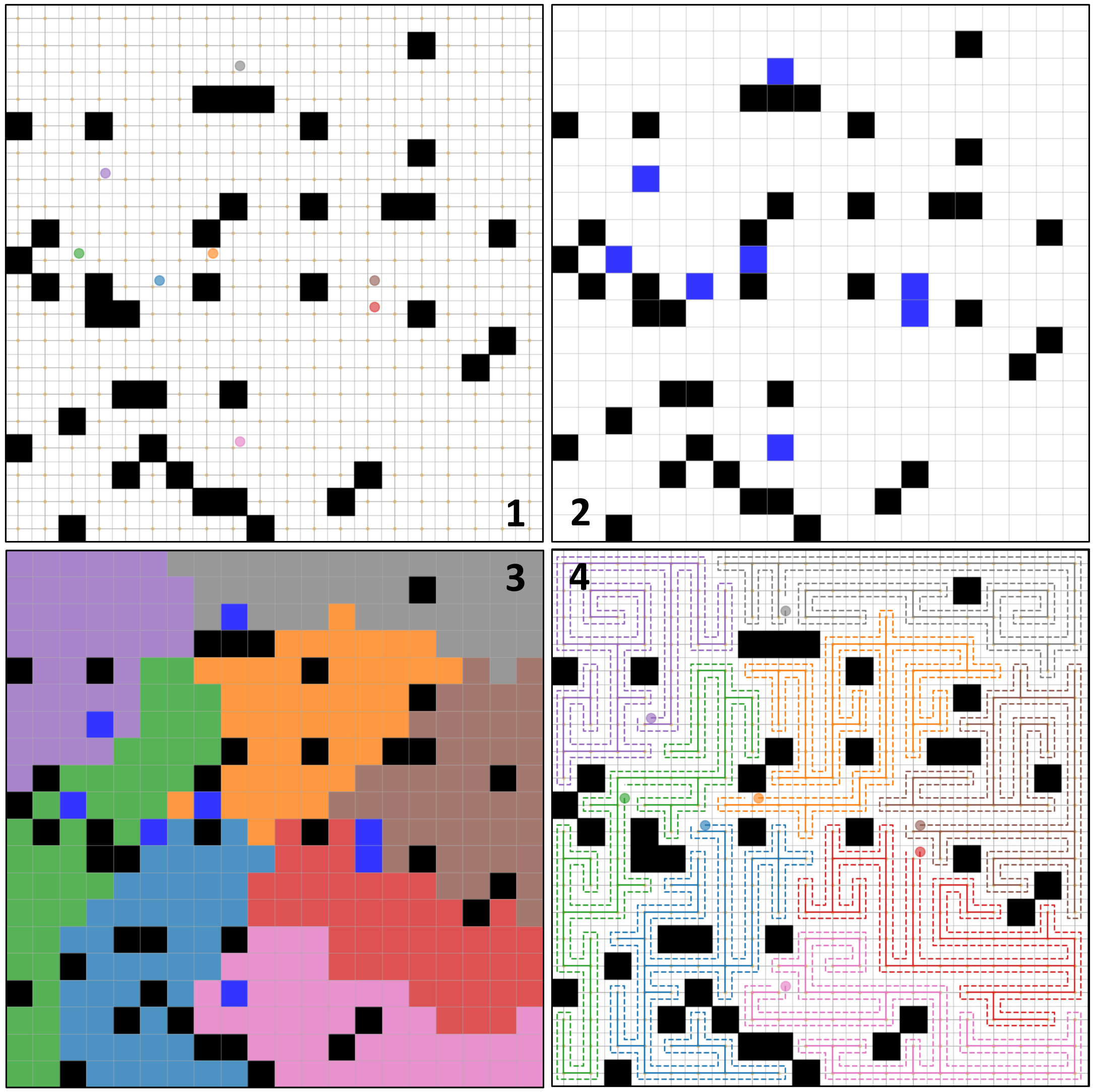}
\caption{An example of solving an mCPP task which can be described in the following order (1) Representing the original map with grids. Each dot in the map indicates an individual robot. (2) Merging each four neighboring grids into cells for STC-based algorithms. A blue cell indicates that a robot exists within the cell.(3) Allocating subareas for each robot. (4) Generating routes based on STC principles.}
\label{fig:mCPP example}
\end{figure}

\section{Problem Description} \label{sec:problem description}
The primary goal of decentralized mCPP is to provide a strategy for each robot to generate a route resulting in full coverage and high efficiency. As our proposed method is based on STC, the interested area \(A\) needs firstly to be decomposed to equal cells turning the area into a graph \(G=(V,E)\), where \(V\) = \(\{n_{1}, ..., n_{r\times c}\}\) is a set of nodes (cell centers) given the width and the length of AOI are respectively split into \(r\) and \(c\) segments, and \(E\) is a set of edges linked only within adjacent cells. Since there exist obstacles \(O\) in area \(A\), we further assume all obstacles fully occupied their cells, which helps us focus mainly on subarea allocation.

Based on the aforementioned preprocess, we then can have \(T=\{T_1, ..., T_m\}\) as a set of subareas, where each subarea \(T_i\) is defined as a group of nodes \(\{l_{i0}, ..., l_{iN_i}\}, l_{ik} \in V\), for \(m\) robots executing STC within their corresponding subareas. The overall processes are illustrated in Fig. \ref{fig:mCPP example}. Then, to accomplish the mentioned purpose, the subarea of each robot is considered to be constructed in a way that
\begin{align} \label{eq:problem description}
    &\min\max \quad  ||T_i|| \nonumber \\
    &\phantom{.}\text{s.t.} \quad   \begin{aligned}[t] 
        O \cup T &= V \\
        O \cap T &= \emptyset
    \end{aligned}
\end{align}
where \(||T_i||\) denotes the total number of nodes (\(=N_i\)) included in the subarea of robot \(i\). The equation \ref{eq:problem description} encourages robots to generate equal-size subareas while both no obstacle collision and full coverage are accomplished.

\section{RL problem casting} \label{sec:RL problem casting}
 To solve mCPP as a multi-agent RL problem, we transform the spanning process in \cite{9017928} into a sequential decision-making process. Then we set up all the key elements for our RL framework, including both observation and action space for our agents and the designed reward function. 
\subsection{Sequential decision process}
According to \cite{9017928},  each robot starts to build a tree from its initial node (position) round by round during the process of constructing spanning trees. At every round, each robot can only extend to a node around its tree, and the whole process is terminated when all free nodes are covered by at least one of the robots. In our work, the procedure is transformed into a sequential decision-making process, and we focus only on determining subareas rather than the tree structures of spanning trees. The process then becomes robots consecutively expanding their territories round by round until the AOI is fully covered. That is, a robot needs to broadcast its decision before another one makes its own. As the subareas are allocated round by round, the algorithm can also be adapted to heterogeneous coverage task \cite{8301580} if give different step number to each robot. Note that implicit decision conflicts in a round are also prevented when converting into a sequential making decision process. Algorithm \ref{alg:DODGE} explains the details of the expanding process.

\subsection{Action space}
\begin{figure}[h]
\centering
\includegraphics[width=0.48\textwidth]{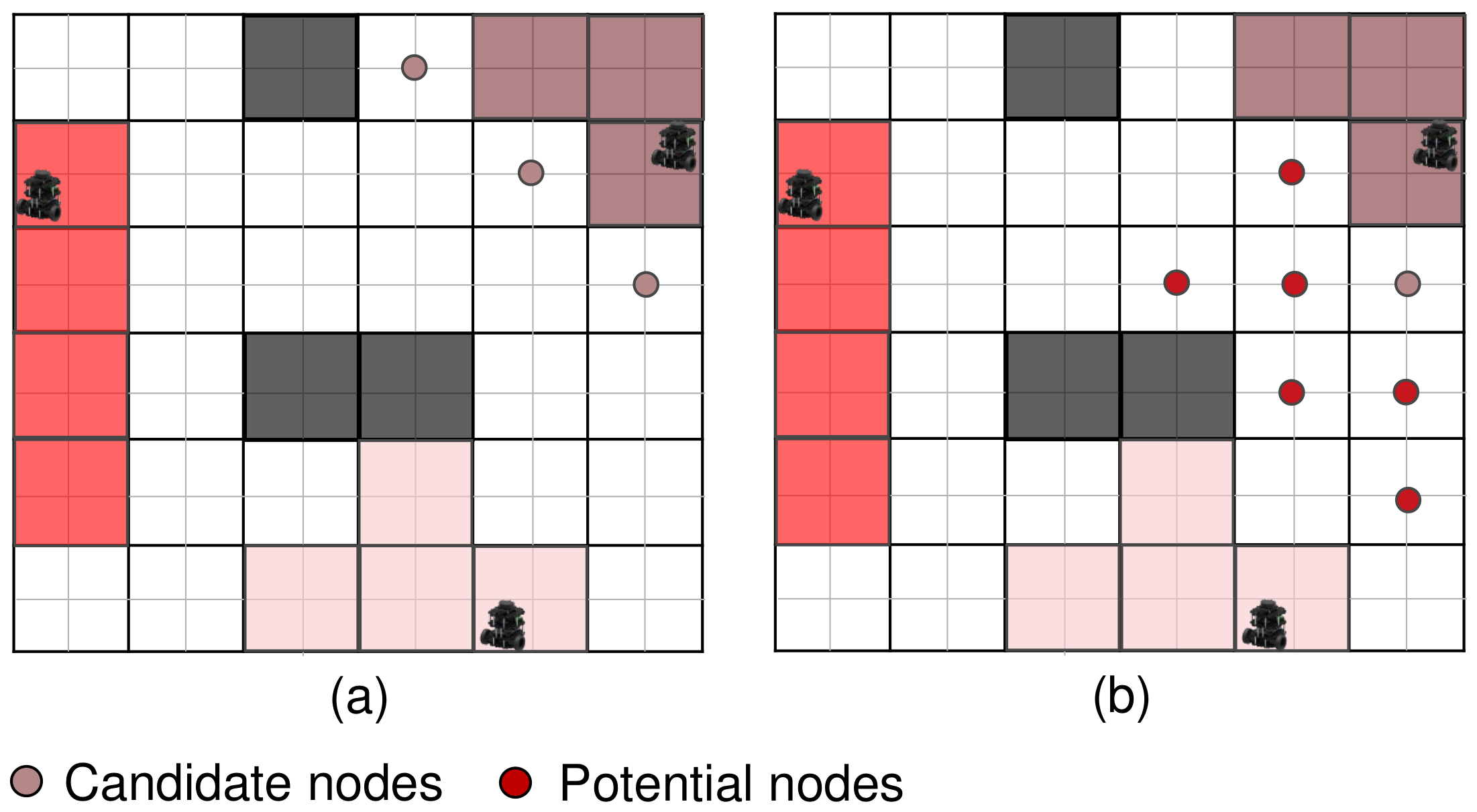}
\caption{(a) Candidate nodes and (b) potential available nodes within next \(f=2\) rounds of the right-up robot.}
\label{fig:hg candidate}
\end{figure}

As described in the introduction, suspending the expanding processes of some agents prevents unnecessary overlaps when a sub-optimal solution comes out. Therefore, we additionally add lazy-evaluation to our action space. Inspired by the use of end token in natural language processing \cite{devlin-etal-2019-bert}, we additionally define a node whose all node features are set to be zero and add it to the node set \(V\) which then becomes \(V_s\). An agent then will not expand its territory if it chooses the lazy-evaluation node as its next move. That is, at round \(t\), each agent has \(A_i^t = [C_i^t, lazy-evaluation]\) as its action space where candidate nodes in \(C_i^t\) are denoted as
\begin{equation}\label{eq2}
    n_j \in C_i^t, \phantom{-}\text{if}\phantom{-} ||(n_j, n_{jT})||_s = 1, \phantom{-}n_j \in V-T_i^t-O
\end{equation}
where \(n_{jT}\) is the nearest node from subarea \(T_i^t\) to node \(n_j\) and \(||(A,B)||_s\) is defined as the shorted path length from node A to node B in Manhattan Distance. An example of candidate nodes is illustrated in Fig. \ref{fig:hg candidate}(a)
.

\subsection{Observation space}
While position information (\(x,y\)) in nodes is not enough for our mCPP tasks, we additionally include 2 features into nodes to aid our agents. In this part, we first describe the added features and our designed node for lazy-evaluation and then introduce the observation of each agent, which consists of three parts: an area state, a subarea state, and a local action mask.

For the first feature, we tend to help agents avoid potential blocks during the expanding process. As agents are likely determining their subareas near around them, we adopt the artificial potential fields \cite{park2001obstacle} to provide distance information \(w\) to node \(n\) for guiding agents to expand their subareas away from others, which is defined as
\begin{equation} \label{eq3}
    w_{n_j}= 
\begin{cases}
    \displaystyle \sum_{i\in\{1,...,m\}}\eta\frac{1}{||(n_j, g_i)||_s+1} & \text{if coverable} \\
    \phantom{------}0         &\text{if obstacles}
\end{cases}
    ,n_j \in V 
\end{equation}

\begin{figure*}[t]
\captionsetup{justification=raggedright,singlelinecheck=false}
\centering
\includegraphics[width=\textwidth]{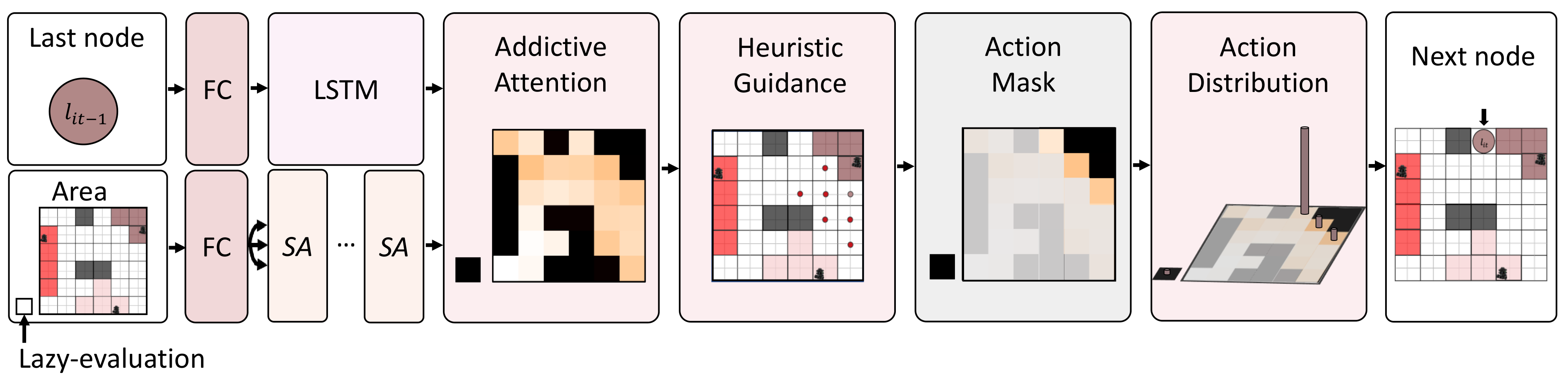}
\caption{Policy architecture. The lighter color in the saliency map means the higher probability the robot is to expand on the node. FC: Fully connected layer; SA: Self attention block.}
\label{fig:policy}
\end{figure*}
where \(\eta\) is a scale factor adjusting \(w_{n_j}\) into [0,1] and \(g_i\) stands for the initial node of agent \(i\). Given the fact that obstacle nodes will not appear in candidates, any value between zero and one should be fine as long as the value is not too large, affecting the learning process of our deep neural network (DNN) policy. For convenience, we set it to be zero for all obstacle nodes.

For the second feature, as agents should know which subarea belongs to which agent, a feature that describes the occupancy information of a cell could help improve agents' performances. To achieve so, we took up the idea of multi-hot encoding. However, the length of a multi-hot encoding vector grows along with the agent number, thus we set a maximum capacity \(C_l\) to the agent number in our method. Considering the fact that decentralized systems often suffer from communication issues, a decent maximum capacity should be enough for most of the tasks. The occupancy information \(o\) of a node is then defined as a zero vector with the length of \(C_l + 1\) where the \(i\)th position of the vector is then marked as one if the cell is occupied by the \(i\)th (agent id) agent and the last position is marked as one if it is occupied by obstacles.

Thus far, we have a set of nodes \(V_s\) presenting the area and lazy-evaluation while each node consists of four features (\(x,y,w,o\)) guiding agents satisfy equation (\ref{eq:problem description}). As most of the works \cite{cao2021dan}, \cite{ma2019combinatorial} have found that relative coordination helps prevent premature convergence of DNN policies and it makes more sense the lazy-evaluation actually points to the initial node of an agent, the position features in nodes are presented in relative coordination. 

Finally, the observation can be illustrated. The observation of agent \(i\) at round \(t\) is then given as the following three components: an area state which contains all the nodes in an agent-specific set \(V_{s_i}^t\) similar to \(V_s\) but (\(x, y\)) are in relative coordination and \(o\) changes over time, an agent-specific subarea state \(T_{s_i}^{t-1}\) which represents all the previous actions including lazy-evaluation and finally a local mask \(M_{i}^{t-1}\) that masks out all the nodes not next to its territory and also the obstacles.

\subsection{Reward design}
Our main purpose is to generate subareas whose sizes are as equal as possible while minimizing overlaps and achieving full coverage. Seeing that one agent's success might lead to a failure of another, it is hard to give credit according to just one agent's performance. We adopt the centralized training decentralized execution framework mentioned in \cite{lowe2017multi} and model our reward function with respect to a team. On top of that, it is also difficult to judge any move for a team during our mCPP tasks, reward is given only at the end of the tasks. A reward function \(R_{m}\) that leads a team of \(m\) robots to fulfill the goal is then designed as:
\begin{equation}\label{eq4}
    R_{m} = -(r_{r} + r_{o})
\end{equation}
where \(r_{r} = ||V - O - T||\) is the number of remaining uncover free nodes and \(r_{o} = \frac{1}{2}\sum_{i,j \in \{1,...,m\}} ||T_i \cap T_j||, i\neq j\) is the number of overlap nodes.

While giving a limited time horizon, a team of agents that maximize the reward function achieves high efficiency (balanced subareas) in covering the interesting area as well as minimizing total redundancy. 

\begin{algorithm}[t]
\caption{Subarea planning process of DODGE}\label{alg:DODGE}
\begin{algorithmic}
\Require $ \text{Area~}V_{s_i}^0, \text{Subarea~} T_{s_i}^0, \text{Action mask~} M_{i}^0, i \in [1,m]$
\State $t \gets 0$
\While{(not full coverage)}
\State $t \gets t + 1$
\For{$i=1,...,m$}
    \State $ V_{s_i}^t \gets \text{Update}\text{(}l_{jt}\text{)}, j \in [1,i-1]$
    \State $ a' \gets \text{Calculate~attention}\text{(}V_{s_i}^t, T_{s_i}^{t-1}\text{)}$ [sec.\ref{sec:RL problem casting}-C] 
    \State $a'' \gets \text{Heuristic~guidance}\text{(}a'\text{)}$ [sec.\ref{sec:RL problem casting}-D] 
    \State $A_i^t \gets \text{Maskout}\text{(}M_{i}^{t-1}, a''\text{)}$  [sec.\ref{sec:problem description}-B] 
    \State $p_{it} \gets \text{Softmax}\text{(}A_i^t\text{)}$
    \State $l_{it} \gets \text{Select}\text{(}p_{it}\text{)}$
    \State $T_{s_i}^t, M_{i}^t \gets \text{Update}\text{(}l_{it}\text{)}$
    \State $\text{Broadcast}\text{(}l_{it}\text{)}$
\EndFor
\EndWhile
\end{algorithmic}
\end{algorithm}

\section{Neural Network Policy} \label{sec:NNpolicy}
For mapping the observation space to the action space, we propose a DNN policy based on long-short term memory (LSTM) and attention mechanism to help each agent interpret both area and subarea states and select the next node. The constructed DNN policy is shown in Fig. \ref{fig:policy} and the details of each element are described in this section.

\subsection{Area encoder}
Since the attention mechanism can be viewed as a weighted information mixer for all given nodes, we encode our area state with attention layers to model the complicated relations among the map and all agents. The area encoder we use is similar to the self-attention layer in Transformer \cite{vaswani2017attention}, including skip connection\cite{he2016deep} and layer normalization\cite{xiong2020layer}, but we only apply single head attention such that the variance is reduced and thus the training process is less unstable. Moreover, before feeding into self-attention layers, node features are embedded into a higher dimension through a linear transformation (a fully connected layer) to capture more information. 

To illustrate our embedding method clearly, given \(h_A = \{h_{A_{1}}, ..., h_{A_{r\times c}}, h_{A_{s}}\}\) as the input embedded vectors of a self-attention layer, which are either the embedded vectors of \(V_{s_i}^t\) from the linear transformation or the embedded vectors from the previous self-attention block. We then transform each vector in \(h_A\) into three vectors with equal dimension \(d\): a query vector \(q\), a key vector \(k\), and a value vector \(v\) which are defined as:
\begin{equation}\label{eq5}
    q_i = W^Qh_{A_{i}}, \phantom{-}k_i = W^Kh_{A_{i}}, \phantom{-}v_i = W^Vh_{A_{i}}
\end{equation}
where \(W^Q, W^K, W^V\) are learnable matrices with size \(d\times d\).
Next, we compute the compatibility \(u_{ij}\) of each two nodes from their query vectors and key vectors using dot-product:
\begin{equation}\label{eq6}
    u_{ij} = (q_i^T\cdot k_j)/\sqrt{d}
\end{equation}
From the above compatibility, we then calculated the attention weights \(a_{ij}\) using a softmax:
\begin{equation}\label{eq7}
    a_{ij} = e^{u_{ij}}/\displaystyle\sum_{j \in V_{si}^t}e^{u_{ij}}
\end{equation}
Afterward, we derive a new vector for each node containing all nodes' information from a weighted sum of their value vectors:
\begin{equation}\label{eq8}
    h_{A_{i}}' = \displaystyle\sum_{j \in V_{si}^t}a_{ij}v_{j}
\end{equation}
Eventually, the derived vectors \(h_{A}'\) are then fed into a feed-forward sublayer to get the final embedding vectors of a self-attention block which consists of a self-attention layer and a feed-forward sublayer. Note that skip connection and layer normalization are implemented after both the feed-forward sublayer and self-attention layer. 

\subsection{Subarea encoder}
In a sense that a representative \(h_{T_{s_i}^{t-1}}\) of a subarea state should focus more on those nodes who have available nodes around them rather than those enveloped inside. Instead of feeding the whole subarea into attention layers, we adopt the LSTM model to utilize the potential ability to both memorize and forget fed-in data. Namely, a subarea is encoded in a way that we feed the output of our policy into our LSTM model round by round, and as like area state, node features are also embedded into a higher dimension \(d\) before feeding into our LSTM encoder.
\subsection{Decoder}
At each round, the decoder gives out the next node based on both embeddings from the subarea encoder and the area encoder. Since the computational cost is less during the decoding phase and the additive attention mechanism \cite{vinyals2015pointer} often shows better results \cite{britz2017massive}, we adopt the addictive attention mechanism and take the embedded subarea state as a query vector and the embedded area state as a reference vector \(r\) to compute attention weight \(a'\) for each node \(n\). Then all nodes' attention weights for agent \(i\) at round \(t\) is calculated as: 
\begin{equation} \label{eq9}
    a_{n_{j}}'= 
    v^T\cdot\text{tanh}(W_rr_{n_j}+W_qh_{T_{s_i}^{t-1}}),
    \phantom{-} n_{j} \in V_{s_i}^t
\end{equation}
where \(v, W_q, W_r\) are all learnable parameters and \(v\) is an one-dimensional vector with size \(d\) and \(W_q, W_r\) are square matrices with size \(d \times d\). The calculated attention weights represent how much the subarea is interested in each node and can later be converted to action distribution using a Softmax operation. Note that we focus more on the whole subarea itself, so we take the cell state (long-term memory) instead of the hidden state (working memory) from the output of the LSTM as a query vector.

\subsection{Heuristic guidance}
To further tackle down the complexity of our mCPP environments, we take into account not only the attention weight of a node in candidate nodes \(C_i^t\) itself but also all its potential nodes \(P_n\), illustrated in Fig. \ref{fig:hg candidate}(b), can be extended within next \(f\) rounds. Then the attention of candidate nodes are modified as
\begin{equation}\label{eq10}
    a_{n_j}'' = \gamma a_{n_j}' + (1-\gamma)\frac{1}{|P_{n_j}|}\sum_{n_k \in P_{n_j}} a_{n_k}'
    ,\phantom{-} n_j \in C_i^t
\end{equation}
where \(\gamma\) is a re-scale factor for the attention weight of a candidate node and \(|P_n|\) denotes the total count of a candidate's potential nodes. Note that we do not implement heuristic guidance on the lazy-evaluation, so the attention weight of a robot's lazy-evaluation action remains the same.

Afterward, we mask out all the unreachable nodes and do a Softmax operation on remaining nodes (candidates and lazy-evaluation) and get the final action distribution which the action distribution for agent \(i\) at each round is denoted as
\begin{equation}\label{eq11}
    p_{it} = \pi_{\theta_i}(l_{it}|V_{s_i}^t, T_{s_i}^{t-1}, M_i^{t-1}) = \text{Softmax}(a_{A_i^t}'')
\end{equation}
where \(\pi_{\theta_i}\) is our constructed policy for agent \(i\) and is parameterized with \(\theta_i\).

\section{Training}\label{sec:training}
In this section, we first introduce the details of our training environments then describe how we train our policy.
\subsection{Environments}
As mentioned in Section-\ref{sec:introduction}, we focus on the outdoor-like environments as introduced in \cite{1545323} where 10\% of the terrain is occupied by obstacles. For our algorithm to work, a terrain is then turned into a graph consisting of \(r\times c\) nodes. Considering the training efficiency, we train our policy in small maps where we set \(r = c \) and \(r, c \in \{9,10\}\). Moreover, the number of agents in the scene is set in a range from 3 to 8 to further alleviate our training costs. In order to extend our policy to the maximum agent capacity \(C_l\), which we set to 20 in our case, the id of an agent is randomly assigned during the training phase. Therefore, our policy can learn the full meaning of feature \(o\) without actually training on larger team sizes. 
\subsection{Parameter sharing}
Due to the homogeneous property among agents in our mCPP task, we apply parameter sharing \cite{gupta2017cooperative}, which makes \(\theta_1=\theta_2,...,=\theta_m=\theta\), to our learning framework allowing all agents to take actions using the same single policy. With parameter sharing, only one policy \(\pi_\theta\) is then left to train, and the whole system can be extended to any agent number that is smaller than our assigned maximum capacity \(C_l\) without training another policy. 
\subsection{Training algorithm}
As the proposed reward function works only if a proper limited horizon \(N\) is given, for every training scenario we set a horizon \(N\) equals to an idealized subarea size: the number of all coverable nodes, excluding the initial nodes, divided by the number of agents. Given the assigned horizon \(N\), a team finds an optimal solution, if it exists, when the reward function is maximized. Accordingly, with the centralized training framework mentioned in Section-\ref{sec:RL problem casting},  we define our policy loss as:
\begin{equation}\label{eq12}
    L_i(\pi_{\theta}) = -\mathbb{E}_{\pi_{\theta}(l_i)}  [R_{m}(\pi_{\theta})]
\end{equation}
where \(\pi_\theta(l_i) = \displaystyle\prod_{t=1}^{N}\pi_\theta(l_{it}|V_{s_i}^t, T_{s_i}^{t-1}, M_i^{t-1})\). 

With loss defined, we optimize the loss by gradient decent using the REINFORCE algorithm \cite{williams1992simple} with a simple exponential moving average baseline \(b\). Giving a decay rate \(\beta\), our baseline is implemented as: b equals \(-L_i(\pi_{\theta})\) in the first iteration then get updated as \(b \gets \beta b - (1-\beta) L_i(\pi_{\theta}) \) in the following iterations. The benefit of adding a baseline into the gradient estimator is that it reduces gradient variance and therefore speeds up the training progress. The policy gradient of our policy is derived as:
\begin{equation}\label{eq13}
    \nabla_{\theta} L_i(\theta)= -\mathbb{E}_{\pi_{\theta}}
    [(R_{m}(\pi_{\theta})-b)\nabla_{\theta}log\pi_\theta (l_i) ]
\end{equation}
If the agent number is less than five, we perform gradient update \(\theta \gets \theta - \alpha \nabla_\theta L_i(\theta)\) for each agent after a batch of instances are finished. Otherwise, if the number of agents is larger than five, we only update through half of the agents' experiences to prevent over-updating our policy in one batch of scenarios.
\subsection{Curriculum learning}
Considering that performance degrades along with the number of agents increases, our exponential moving average baseline fluctuates if we randomly set the number of agents, which consequently leads to high variance in policy gradient and thus slow and unstable learning. To overcome the above issue, we first train our policy for scenarios that consist of only three agents and incrementally include more agents into tasks as described in \cite{gupta2017cooperative}, but instead of doubling our agent number, we add only one agent at a time when a team reaches assigned performance threshold. 
\begin{table*}[t]
\captionsetup{justification=centering, labelsep=newline}
\caption{Cover time for 98x98 outdoor-like terrains(in terms of path length)}
\begin{tabular}{ p{2.095cm}p{2.095cm}p{2.095cm}p{2.095cm}p{2.095cm}p{2.095cm}p{2.095cm}p{0cm}}
 \hline
  & \centering \textbf{Agents} & \centering \textbf{Ideal Max} & \centering \textbf{Max} & \centering \textbf{Min} & \centering \textbf{Ratio} & \centering \textbf{Overlap rate} &\\
 \hline
  DARP   & \centering 8 & \centering 1079 & \centering 1082 
 & \centering 1078 & \centering 1.003  & \centering - &\\
 
 MFC   & \centering 8 & \centering 1079 & \centering 1247 
 & \centering 873 & \centering 1.16  & \centering - &\\
 
 AWSTC   & \centering 8 & \centering 1032 & \centering 1080 
 & \centering 1076 & \centering 1.047  & \centering \textless4.66\%&\\
 
 \textbf{DODGE (ours)} & \centering 8 & \centering 1081 & \centering 1216 
 & \centering 880 & \centering 1.13  & \centering \textbf{\textless0.02\%} &\\
 \hline
 DARP   & \centering 14 & \centering 616 & \centering 620 
 & \centering 616& \centering 1.006  & \centering - &\\
 
 MFC   & \centering 14 & \centering 616 & \centering 746 
 & \centering 464 & \centering 1.21  & \centering - &\\
 
 AWSTC   & \centering 14 & \centering 600 & \centering 644 
 & \centering 640 & \centering 1.073  & \centering \textless7.33\%&\\
 
 \textbf{DODGE (ours)} & \centering 14 & \centering 618 & \centering 684 
 & \centering 460 & \centering 1.11  & \centering \textbf{\textless0.03\%} &\\
 \hline
 
 DARP   & \centering 20 & \centering 431 & \centering 434
 & \centering 430 & \centering 1.007  & \centering - &\\
 
 MFC   & \centering 20 & \centering 431 & \centering 551 
 & \centering 296 & \centering 1.28  & \centering - &\\
 
 AWSTC   & \centering 20 & \centering 420 & \centering 464 
 & \centering 460 & \centering 1.1045  & \centering \textless10.47\% &\\
 
 \textbf{DODGE (ours)} & \centering 20 & \centering 432 & \centering 496 
 & \centering 284 & \centering 1.15  & \centering \textbf{\textless0.09\%}&\\
 \hline\\
\end{tabular}
\label{tab1}
\end{table*}

\section{Experiments}\label{sec6}
In our experiments, we first verify that our proposed elements truly guide agents to construct more efficient routes. Next, we verify our policy can extend to larger maps and more agents and then compare the performance with other existing methods. Lastly, we demonstrate the ability to deal with robot failure and on-board experiments. For all experiments, we take \(f=2, \gamma =0.2\) for the heuristic guidance, vector length \(d=64\) for all embedded vectors, and three self-attention blocks for area encoder.

\subsection{Ablation experiments}
\begin{figure}[ht]
\centering
\captionsetup{justification=raggedright,singlelinecheck=false}
\includegraphics[width=0.485\textwidth]{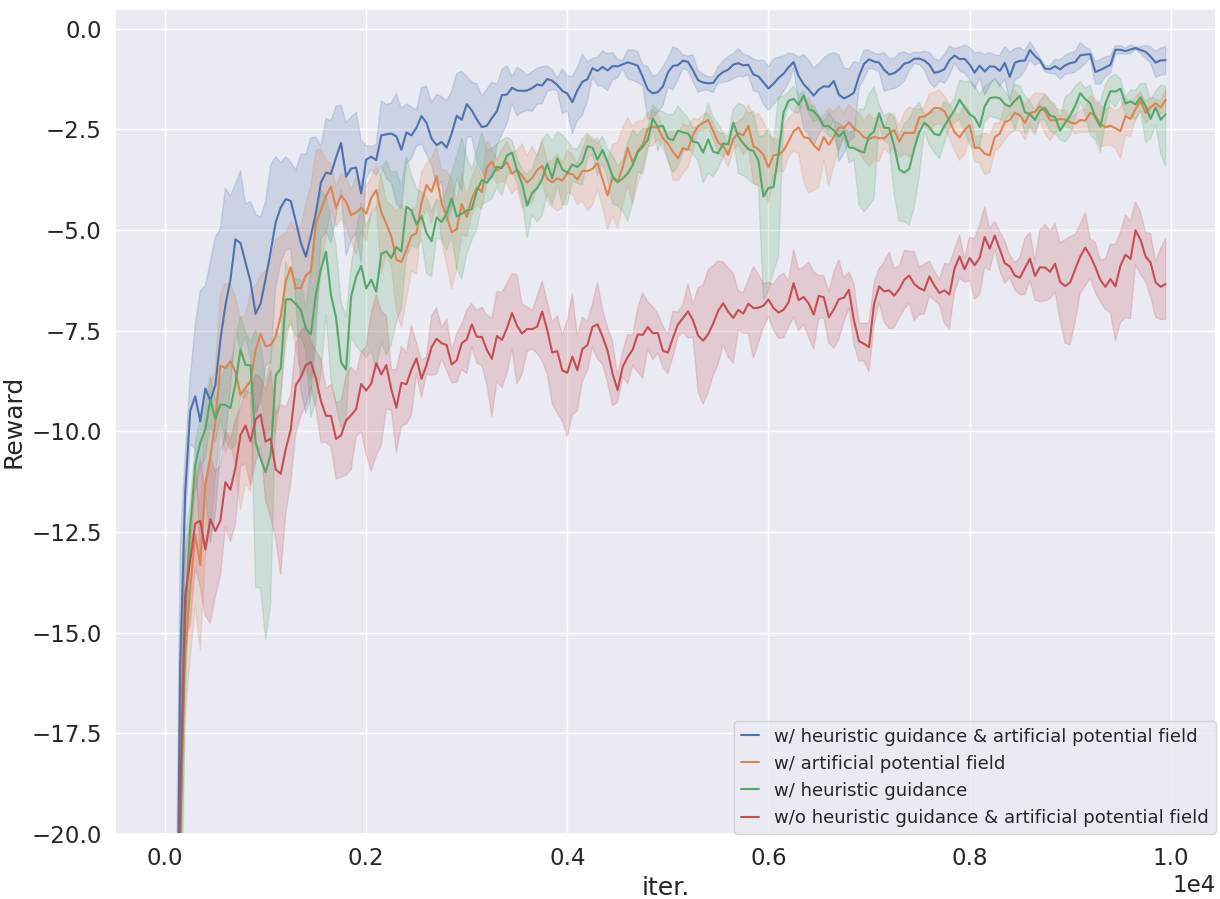}
\caption{Ablation results of different training settings where optimal solution appears when reward equals 0.}
\label{fig:ablation}
\end{figure}

To verify our guiding approaches improve the overall performance, we train our algorithm with the following four settings: our original implementation, with only artificial potential fields feature, with only heuristic guidance, and without both elements. To exclude other factors, we train all settings in \(20\times 20\) outdoor-like terrains with four agents and no curriculum learning is applied. The results are shown in Fig. \ref{fig:ablation} presenting the team reward. While optimal solution equals \(R_m =0\), we observe that both approaches improve the team reward and near-optimal solution yielded with both artificial potential fields feature and heuristic guidance. Thus, the results have proven that both our guidance help agents interpret the interested region resulting in more balanced subareas and lower overlaps.

\begin{figure}[ht]
\centering
\includegraphics[width=0.485\textwidth]{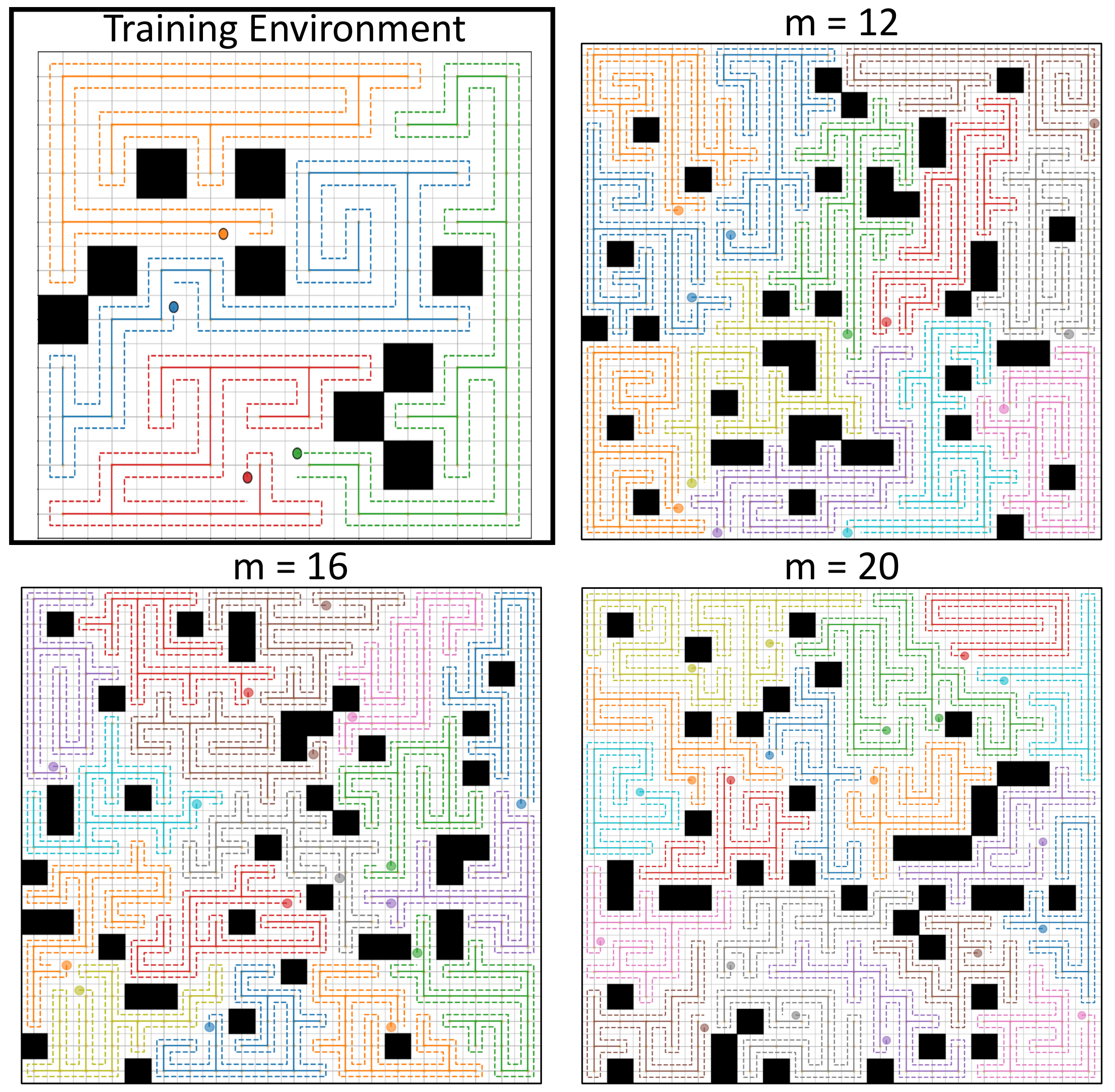}
\caption{Examples of our training environments (20x20) and the capability of extending to larger maps (40x40) and more robots (12, 16, 20).}
\label{fig:extension examples}
\end{figure}

\subsection{Performance experiments}
As current reinforcement approaches \cite{9330612},\cite{8963663} for mCPP problems rely on a central planner or Q-value tables. Existing policies are hard to extend to more agents or larger maps. On the other hand, as our policy is based on neural networks and the decentralized framework, it is capable of extending to both more agents and larger maps. To verify, we implement our policy in larger 40x40 maps with more agents (12, 16, and 20), which are different from our small training scenarios described in Section-\ref{sec:training}. Results are shown in Fig. \ref{fig:extension examples}, our policy successfully determines the subareas and the routes of each scenario are sufficiently equal and without overlaps. To further analyze the performance of our policy, we then compare our performance with a number of existing algorithms and report its statistical results.  

For comparison, as experiments in \cite{kapoutsisdarp}, we implement our trained policy in \(98\times 98\) outdoor-like terrains with three scenarios, where agent number varies from 8, 14, to 20, and the initial position of agents are randomly assigned. In order to analytically evaluate the effectiveness of our method, we run each scenario 100 times and report the \textquote{Max}, \textquote{Min}, \textquote{Ideal Max}, \textquote{Ratio} and \textquote{Overlap rate} of our experiments as shown in Table \ref{tab1}. \textquote{Max} and \textquote{Min} are the maximum and the minimum of coverage paths length, and \textquote{Ideal Max} represents the optimal length of an agent which is the number of coverable mini-cells divided by the number of agents. \textquote{Ratio}, which is defined as \textquote{Max} divided by \textquote{Ideal Max}, is an index for quantifying the imbalance of resulting routes. The \textquote{ratio} increases when the difference between agents' route lengths becomes larger. As optimal solutions in our scenarios should include no overlaps, we also show \textquote{Overlap rate}, which is defined as overlap cell number divides coverable cell number.

Next, we compare our results with a number of algorithms that also tested their performance in similar scenarios, which include two centralized methods, DARP \cite{kapoutsisdarp} and MFC, and one decentralized method, AWSTC. As it is hard to optimally solve an mCPP problem while lacking access to all agents, our method shows larger \textquote{ratio}s than DARP, which is the first centralized algorithm that equally divides the interested area. However, our method performs lower \textquote{ratio}s than MFC as AWSTC does but remains comparable low overlap rates compared to AWSTC. Results have proven that our framework has successfully trained our agents to allocate sufficiently balanced subareas, and at the same time, reduce unnecessary overlap areas. 

\begin{figure}[ht]
\centering
\subfloat[]{\includegraphics[width=0.485\textwidth]{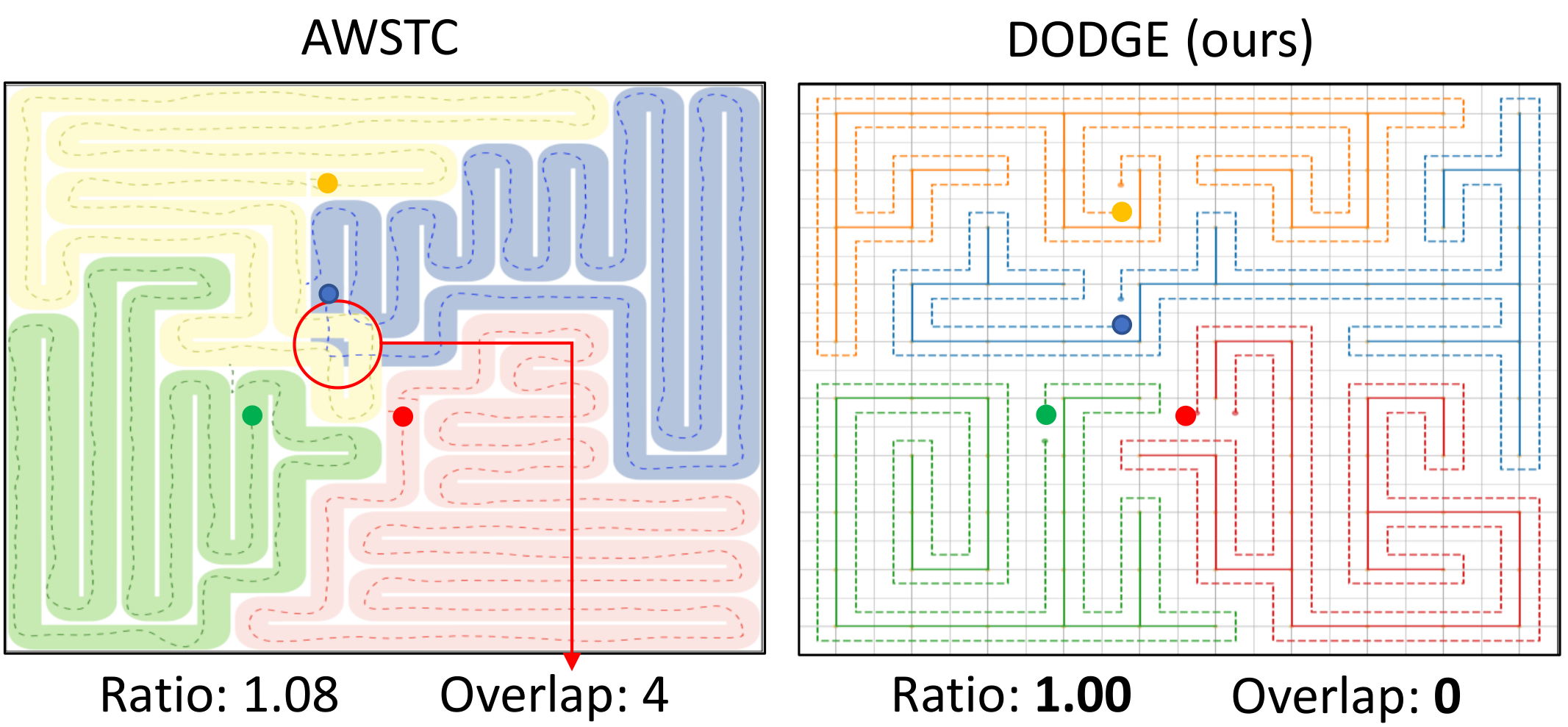}} 
\hfill
\subfloat[]{\includegraphics[width=0.485\textwidth]{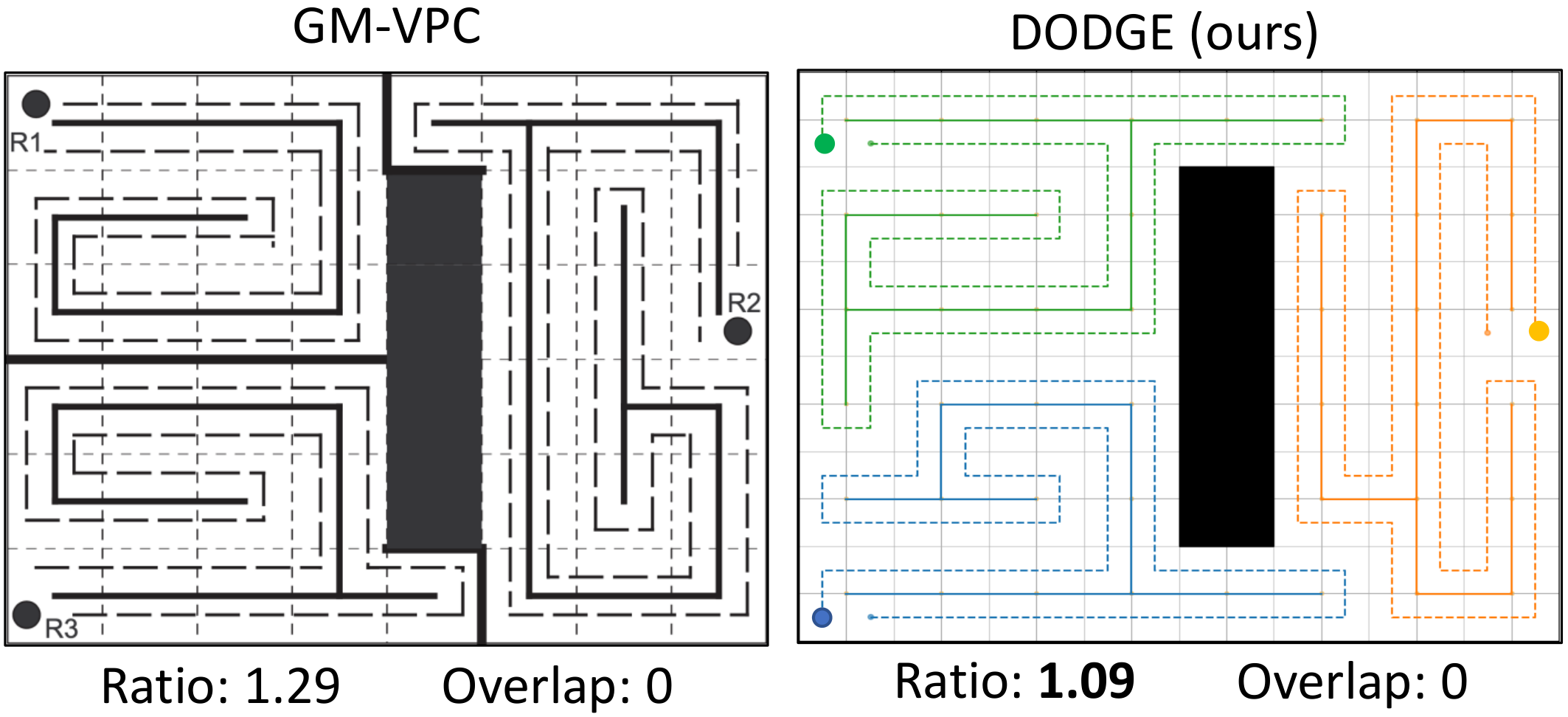}}
\captionsetup{justification=raggedright,singlelinecheck=false}
\caption{Comparative results against existing decentralized methods. (a) with AWSTC (b) with GM-VPC.}
\label{fig:compare existing decentralized methods}
\end{figure}

To show the benefit of our algorithm, we also compared the generated routes with the existing decentralized methods, AWSTC and GM-VPC, in the same scenario proposed in their articles. Despite that our \textquote{ratio}s are higher than AWSTC in the previous large map test, our result in Fig. \ref{fig:compare existing decentralized methods}(a) has shown a lower \textquote{ratio} while remaining lower overlaps within the same small scenario proposed in the AWSTC article. Moreover, as GM-VPC simply relies on the distance relationship of robot position, our result in Fig. \ref{fig:compare existing decentralized methods}(b) also shows more balanced routes within the same scenario. That is, our algorithm has the potential to perform better than existing methods in scenarios that they fail to optimize.

\begin{figure*}[t]
  \centering
  \subfloat[]{\includegraphics[width=0.48\textwidth]{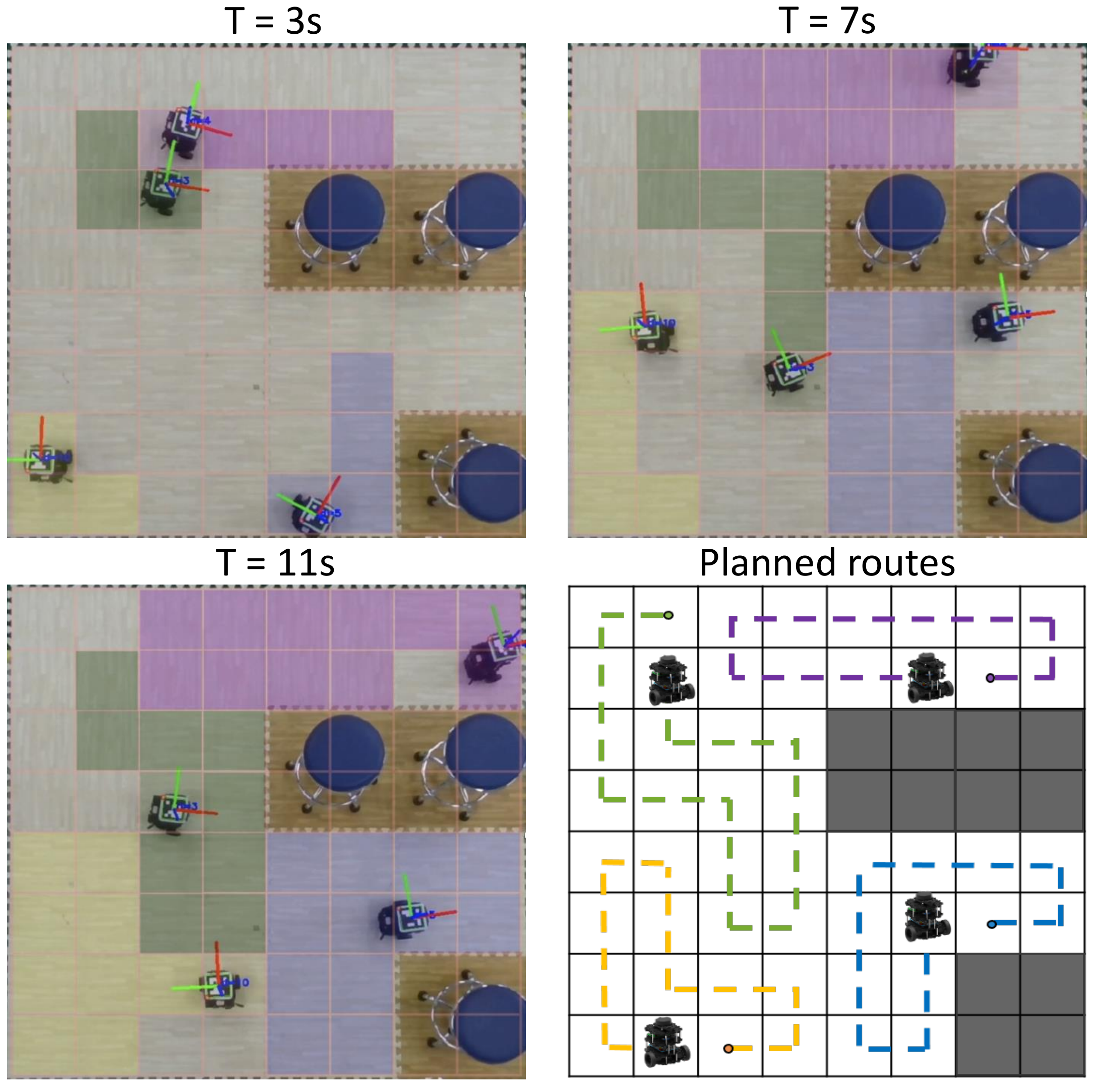}}
  \hfill
  \subfloat[]{\includegraphics[width=0.48\textwidth]{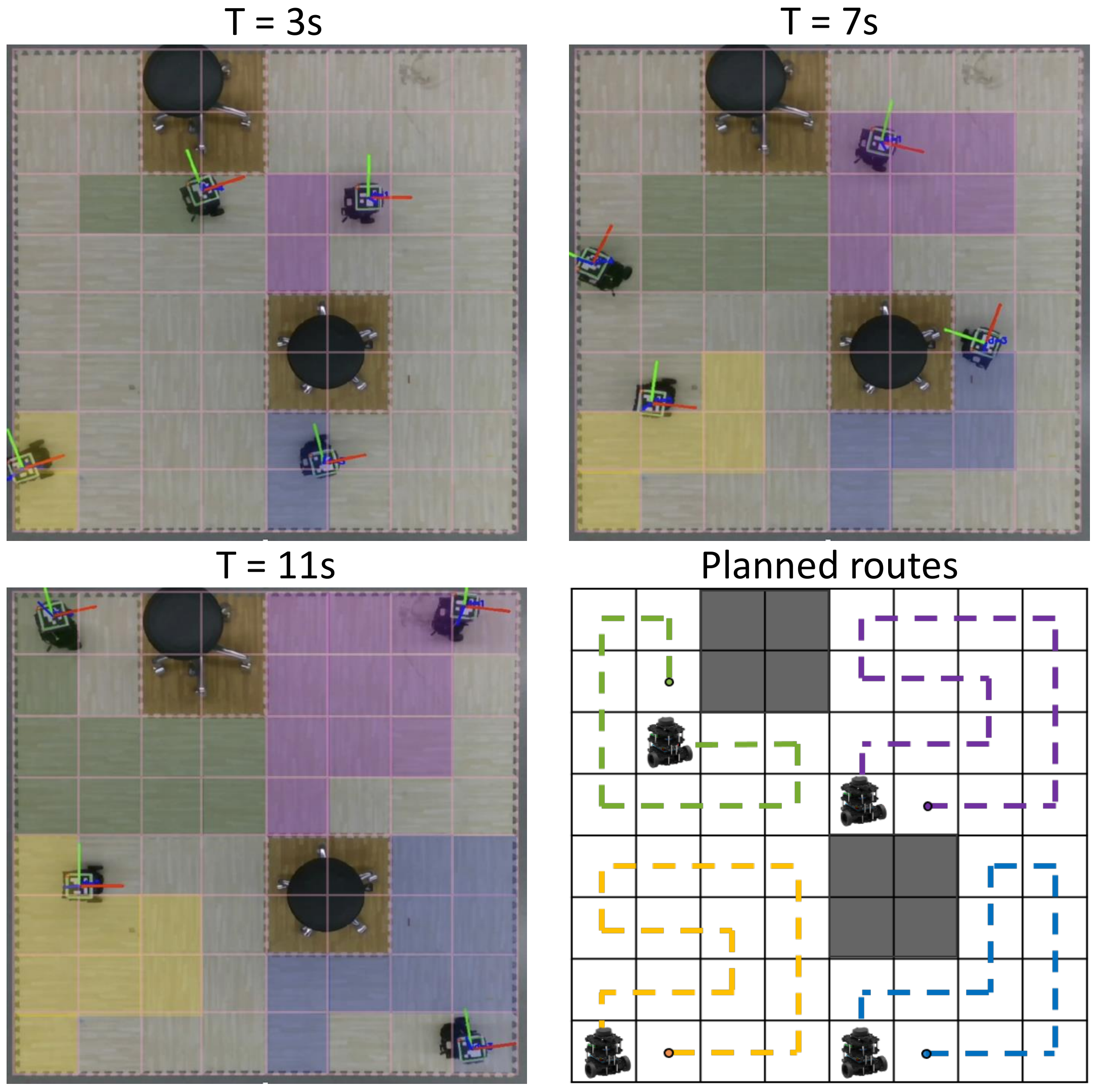}}
  \captionsetup{justification=raggedright,singlelinecheck=false}
  \caption{On-board experiments in different 8x8 areas with four robots.}
  \label{fig:On-board experiments in different 8x8 areas with four robots}
\end{figure*}

\begin{figure}[ht]
\centering
\subfloat[]{\includegraphics[width=0.485\textwidth]{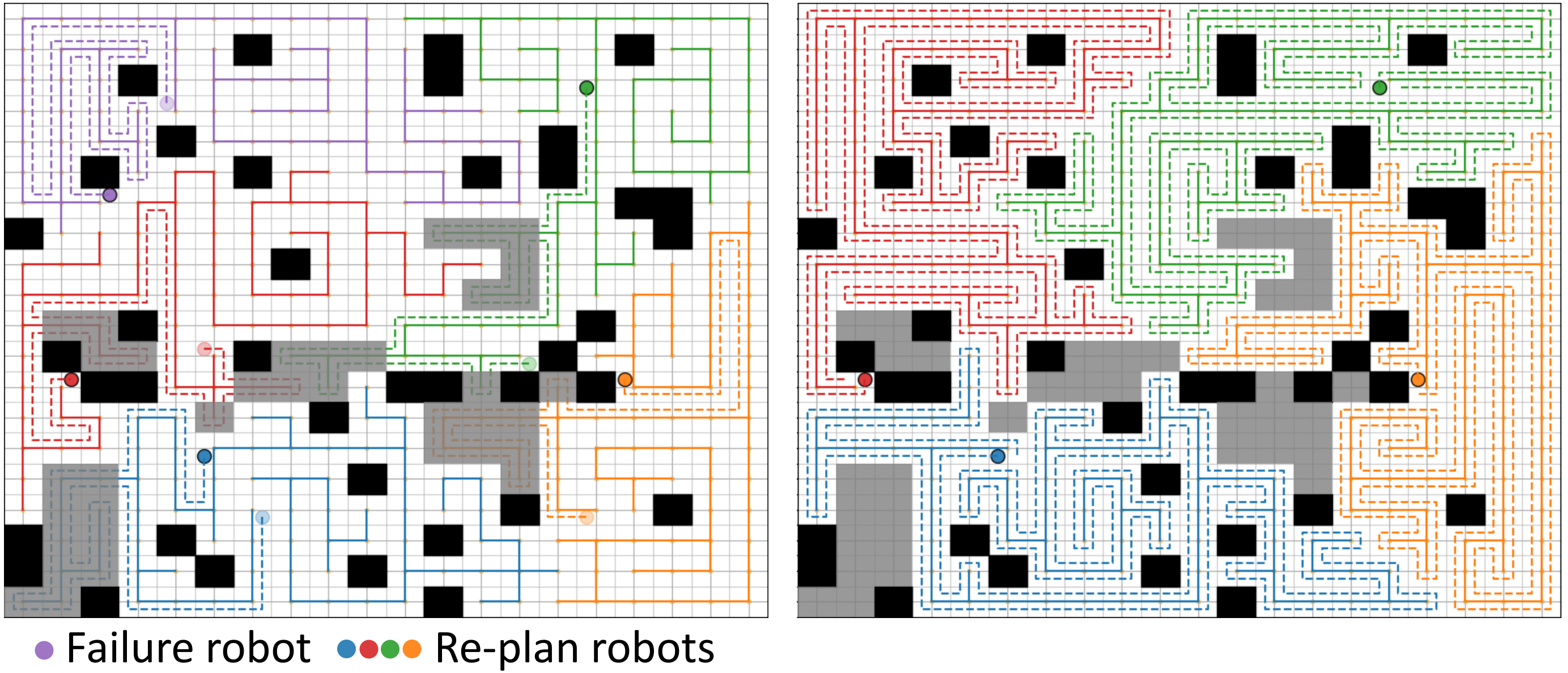}} 
\hfill
\subfloat[]{\includegraphics[width=0.485\textwidth]{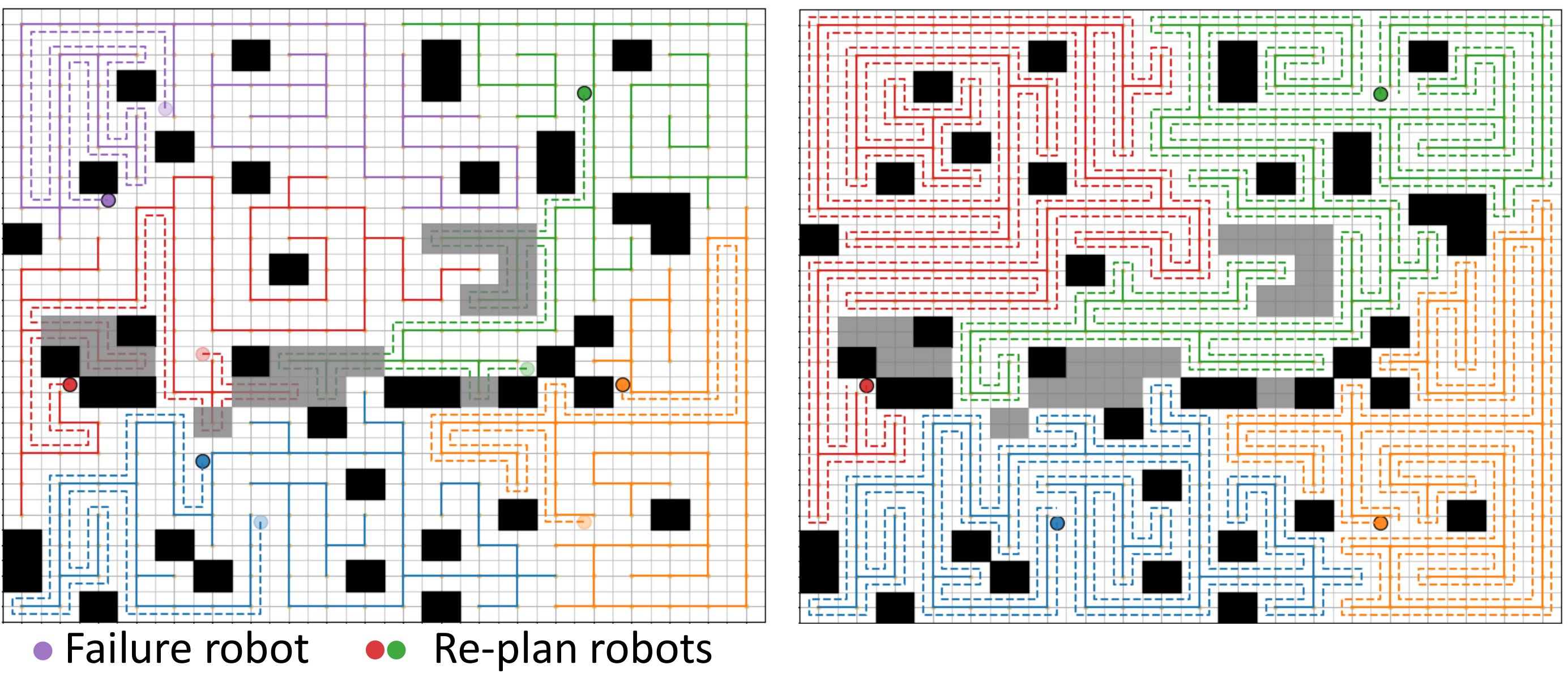}}
\captionsetup{justification=raggedright,singlelinecheck=false}
\caption{Subarea reallocation for dealing with robot failure using (a) the rest of robots (b) neighbor robots.}
\label{fig:failure replanner}
\end{figure}

\subsection{Failure re-planner}
 During the task, robots might not be able to reconnect to the central system due to the communication constraints. When the situation occurs, incomplete coverage happens when some of the robots break down in the mid way. On the other hand, as a decentralized method, our framework can completes the coverage even a robot breaks down as shown in Fig. \ref{fig:failure replanner}. While a robot (purple) is broken, we can utilize either the rest of the robots or only its neighbor robots to reallocate the subarea of the broken robot without connecting to a central device. In practice, we ignore the cells (grey) which are already fully covered by the alive robots who participated in the reconstructing phase before a broken robot is detected, then set the remaining cells free and re-plan the coverage path for those robots. As shown in Fig. \ref{fig:failure replanner}, results have demonstrated our algorithms are capable of handling such failure mode while centralized methods usually cannot.

\subsection{On-board experiments}
To further test our policy's generality and efficiency, we present results of on-board experiments implemented using Turtlebot3 ground vehicles with a pseudo-decentralized system. We conduct our planned routes on 8x8 area with 4 robots in different environments as shown in Fig. \ref{fig:On-board experiments in different 8x8 areas with four robots}. The results have endorsed our claimed performance that our policy allocates balanced subareas and few overlap areas are introduced.

\section{Conclusions}\label{sec7}
In this work, we present a deep reinforcement learning framework giving informative guidance for training a team of robots to solve mCPP problems in a decentralized manner. Each robot learns to intelligently expand its territory from its initial position within the interested region round by round until the whole free area is covered by at least one of the robots. With our proposed guiding methods, results show our DRL-based STC method successfully achieves full coverage and sufficiently distributes agents' burdens while reducing unnecessary overlaps. Furthermore, despite the fact that our policy is trained in small area sizes and agent numbers, we also verify that our policy can extend to larger maps and more agents while preserving decent performance. Note that in the experiments our policy is trained in outdoor-like environments where no overlaps exist in optimal solutions, the learned strategy of our learner policy inclines to implement lazy-evaluation while all its territory candidates are covered by others' territories. Our policy theoretically can learn to solve any given known environment if it is trained on different and various scenarios like corridors, offices, and indoor-like terrains. In the future, we will investigate exploiting our framework for more scenarios and better interpreting the whole inter-relation within the concerned area and agents' interactions in our policy structure.



\end{document}